\documentclass{svproc}
\usepackage{url}
\usepackage{graphicx}

\usepackage{rotating}
\usepackage{amssymb}
\usepackage{mathtools}
\usepackage[table]{xcolor} 
\usepackage{algorithm,algpseudocode}
\usepackage{float}
\begin{document}
\mainmatter              
\title{Text Detoxification in isiXhosa and Yorùbá: A Cross-Lingual Machine Learning Approach for Low-Resource African Languages}
\titlerunning{Text Detoxification in isiXhosa and Yorùbá}  
%
\author{Abayomi O. Agbeyangi}
\authorrunning{Agbeyangi} 
%
\tocauthor{Abayomi O. Agbeyangi}
\institute{Walter Sisulu University, Buffalo City Campus, East London, South Africa,\\
\email{aagbeyangi@wsu.ac.za}}

\maketitle              

\begin{abstract}
	Toxic language is one of the major barrier to safe online participation, yet robust mitigation tools are scarce for African languages. This study addresses this critical gap by investigating automatic text detoxification (toxic to neutral rewriting) for two low-resource African languages, \textit{isiXhosa} and \textit{Yorùbá}. The work contributes a novel, pragmatic hybrid methodology: a lightweight, interpretable TF–IDF + Logistic Regression model for transparent toxicity detection, and a controlled lexicon- and token-guided rewriting component. A parallel corpus of toxic to neutral rewrites, which captures idiomatic usage, diacritics, and code-switching, was developed to train and evaluate the model. The detection component achieved stratified K-fold accuracies of 61–72\% (isiXhosa) and 72–86\% (Yorùbá), with per-language ROC-AUCs up to 0.88. The rewriting component successfully detoxified all detected toxic sentences while preserving $100\%$ of non-toxic sentences. These results demonstrate that scalable, interpretable machine learning detectors combined with rule-based edits offer a competitive and resource-efficient solution for culturally adaptive safety tooling, setting a new benchmark for low-resource Text Style Transfer (TST) in African languages.
	
	
\end{abstract}

\section{Introduction}
As digital platforms increasingly mediate human interaction, the prevalence of toxic language, including insults, threats, and culturally insensitive remarks, presents a growing challenge to safe and inclusive online spaces \cite{Arora2023,SHETH2022}. While considerable progress has been made in detecting and mitigating toxic content using natural language processing (NLP) techniques \cite{sourabrata2023,YI2025128684}, these advancements have primarily focused on high-resource languages, such as English, leaving a significant gap in tools and resources for African languages. The scarcity of annotated datasets, combined with cultural and linguistic diversity, complicates the effectiveness and applicability of existing models in low-resource contexts. 

\begin{figure}
	\centering
	\includegraphics[width=4in]{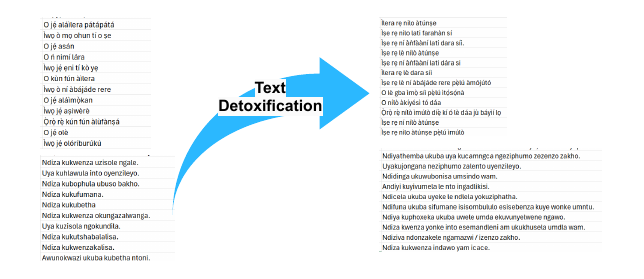} 
	\caption{Text detoxification sample.}
	\label{textDetox}
\end{figure}

Text detoxification (often referred to as text style transfer (TST)) is a method for transforming toxic or offensive text (example in Figure \ref{textDetox}) into a more neutral or respectful form while preserving its original meaning and intent \cite{Dementieva2021,floto2023,Mukherjee2024}. Most approaches typically involved identifying and removing toxic words based on predefined vocabularies. With recent advances in neural models and large-scale pretraining, the quality of detoxification outputs has significantly improved, enabling more context-aware and semantically correct rewriting \cite{CHAN2024,Dementieva2023}. Conversely, most of this progress has been concentrated even in high-resource languages. Consequently, low-resource languages, particularly from Africa, remain underserved in this domain.

isiXhosa and Yorùbá, spoken widely in South Africa and Nigeria, respectively, are linguistically rich and culturally significant African languages that remain underrepresented in the field of natural language processing (NLP) \cite{mesham2021,Agbeyangi2024,UGWU2024}. Despite their widespread use, isiXhosa and Yorùbá face persistent challenges, such as the scarcity of annotated datasets, limited integration into mainstream multilingual models, and minimal representation in existing research efforts. Specifically, based on available literature, the task of text detoxification (Text Style Transfer or TST) remains largely unexplored for isiXhosa and Yorùbá. Foundational work, such as AfriHate \cite{Muhammad2025AfriHate}, has provided crucial datasets for toxicity detection in these languages. Advancements in multilingual NLP, particularly the PAN TextDetox challenge\footnote{https://pan.webis.de/clef25/pan25-web/text-detoxification.html}, which established parallel detoxification data for Amharic \cite{Dementieva2024Multilingual,dementieva2024multipara}, provide promising foundations; no known work has previously addressed end-to-end, meaning-preserving detoxification rewriting for isiXhosa and Yorùbá. Notably, for isiXhosa, cross-lingual transfer techniques \cite{Louw2024} and multilingual model adaptation \cite{Gillis-Webber2018} offer a viable strategy for extending the capabilities of detoxification. For Yorùbá, foundational work in machine translation efforts \cite{Adewole2020,Akinwale2015,Agbeyangi2017}, text-to-speech synthesis \cite{Afolabi2013,Babatunde2024,Iyanda2014}, text synthesis \cite{Aoga2016,Agbeyangi2020}, and corpus development \ cite {Akinwale2015, Agbeyangi2017} lays the foundation for implementing more advanced tasks, such as text detoxification.

While foundational work has introduced some datasets for toxicity detection in multiple languages, and dedicated parallel corpora for detoxification are emerging for languages like Amharic, a functional detoxification system for isiXhosa and Yorùbá has been critically missing. This study addresses this gap by presenting the first dedicated end-to-end detoxification approach for both languages. The study presents a novel, pragmatic hybrid methodology that utilises a lightweight, interpretable TF–IDF + Logistic Regression model for transparent toxicity detection, and employs a controlled lexicon- and token-guided rewriting component. This departure from resource-intensive, black-box large language models (LLMs) offers a computationally efficient and culturally adaptive solution that reliably preserves diacritics and idiomatic usage, setting a new benchmark for low-resource NLP safety tooling and motivating future research into controlled, nuance-sensitive cross-lingual text style transfer.

\section{Related Work}
Text detoxification, as a subtask of text style transfer (TST), involves modifying the stylistic properties of a sentence, such as tone, sentiment, or toxicity, while preserving its semantic content \cite{Dementieva2021,Mukherjee2024,Dementieva2023}. It focuses explicitly on rewriting offensive or toxic text into a more neutral or non-offensive form \cite{floto2023}. Most of the early research in TST utilised rule-based methods \cite{sourabrata2023,CHAN2024,dementieva2024,logacheva2022} and handcrafted features; however, the field has since advanced with the development of neural models and large-scale pretraining. According to Logacheva et al. \cite{logacheva2022}, rule-based detoxification methods, such as the Delete model, remove toxic words using a predefined vocabulary, effectively censoring offensive content. These methods produce outputs that are easier to classify for toxicity but may lack nuanced rewriting, as they primarily eliminate toxic tokens rather than paraphrase sentences. Research by Dementieva et al. \cite{dementieva2024} demonstrated that rule-based methods provide a minimal baseline compared to state-of-the-art approaches. Thus, emphasising the importance of the recent advancement in NLP.

State-of-the-art approaches, such as sequence-to-sequence learning \cite{Sutskever2014}, adversarial training \cite{wong2020}, and controlled generation \cite{Zhiting2017,Kumar2020}, have become common for tasks like politeness transfer, sentiment modification, and reducing toxic content. Floto et al. \cite{floto2023} introduced DiffuDetox, a mixed diffusion model for text detoxification, combining a conditional diffusion model that reduces toxicity in text with an unconditional model that improves fluency. The approach addresses challenges from limited detoxification data by generating a diverse set of detoxified sentences with high fluency and content preservation. The performance of the model on the ParaDetox dataset\footnote{https://github.com/s-nlp/paradetox} achieved a J score of 0.67 and also shows improvements in BLEU\footnote{https://huggingface.co/spaces/evaluate-metric/bleu} (62.13 vs 64.53). Similarly, Logacheva et al. \cite{logacheva2022} employed advanced models, such as ruT5\footnote{https://huggingface.co/ai-forever/ruT5-base} and RuGPT3-XL\footnote{https://huggingface.co/ai-forever/rugpt3xl}, alongside baselines that included rule-based methods and fine-tuned large pre-trained language models. The evaluation with human references scored highest manually (joint score Jm = 0.65), closely followed by ruT5-based models (e.g., ruT5-clean Jm = 0.63). Also, Logacheva et al. \cite{logacheva2022para} created parallel datasets (ParaDetox and filtered ParaNMT) for toxic-to-neutral sentence pairs and fine-tuned the BART model, achieving performance superior to existing unsupervised and other baseline methods on both automatic metrics (e.g., BLEU, J) and human evaluations. They all focus on the languages English \cite{floto2023,logacheva2022para} and Russian \cite{logacheva2022}.

Another notable state-of-the-art approach for massively multilingual machine translation was developed by Chan and Li \cite{CHAN2024}. They introduced “Specialis Revelio”, a text pre-processing module that significantly enhances the detection of disguised toxic content by applying steps like typo correction, slang and leetspeak removal, and word-boundary fixes. Experimental results show that integrating Specialis Revelio with toxic detection APIs, such as Detoxify and Perspective API, leads to notably higher confidence and accuracy in identifying toxic content. Detoxify’s toxicity detection probability increased to above 0.95 after pre-processing, compared to 0.8 without it. Similarly, Dementieva et al. \cite{Dementieva2023} explored a cross-lingual style transfer approach focusing on transferring detoxification capabilities between English and Russian. They compared several approaches, including back translation, training data translation, adapter-based methods, and end-to-end simultaneous detoxification and translation models. The evaluations show that the back-translation approach achieves the highest performance but requires multiple inference steps and relies on the availability of the translation system.

Specifically, cross-lingual NLP through multilingual pre-trained models and transfer learning has consistently bridged the resource gap in text detoxification \cite{Dementieva2023,Beniwal2025,Louw2024}. Models such as mBERT, XLM-R, mT5, and Flan-T5 have demonstrated promise in transferring learned representations across languages. For instance, Dementieva et al. \cite{Dementieva2023} explored methods such as adapter-based fine-tuning of multilingual language models, which allow transfer of detoxification knowledge from a resource-rich language (English) to a low-resource language (Russian). Beniwal et al. \cite{Beniwal2025} demonstrated that cross-lingual detoxification using multilingual pre-trained language models effectively reduces toxicity, achieving substantial toxicity reduction even with limited fine-tuning data (10-30\%). Their approach seems beneficial in scenarios where training data is limited, and could be explored in low-resource language settings.

Recently, several NLP competitions have increasingly addressed the task of text detoxification, recognising its importance in promoting safer and more inclusive online communication. Shared tasks such as those hosted by SemEval\footnote[5]{https://semeval.github.io/}, Pan at CLEF\footnote{https://pan.webis.de/clef25/pan25-web/text-detoxification.html}, and emerging initiatives like ParaDetox\footnote{\cite{logacheva2022para}} have challenged researchers to develop models capable of identifying and transforming toxic or offensive text into more neutral or respectful language while preserving the original meaning. For example, Dementieva et al. \cite{dementieva2024} reported the shared challenge of the Multilingual Text Detoxification task at PAN 2024, which involves detoxifying toxic language across nine languages, including English, Spanish, German, Chinese, and Arabic. They noted that participants used fine-tuned or prompted state-of-the-art LLMs like mT0-XL, GPT-3.5, and LLaMa-3, achieving near or above human-level performance in resource-rich European languages (English, Spanish, German). However, performance lagged notably for less-resourced languages such as Chinese, Hindi, and Amharic. Despite strong automatic evaluation results, especially from multilingual models, challenges remain in toxicity handling and consistent cross-lingual transfer \cite{dementieva2024}. Similarly, Dementieva et al. \cite{dementieva2022russe} reported the first Russian detoxification challenge focused on rewriting toxic text into neutral text. The shared tasks demonstrated that models based on fine-tuned ruT5-large pre-trained Transformers achieved the best performance, producing outputs of high quality.

Notably, several studies have also gathered, curated, or developed datasets specifically for text detoxification tasks, providing essential resources for training and evaluating detoxification models. These datasets typically consist of parallel sentence pairs, where toxic inputs are aligned with their non-toxic or neutralised counterparts. For instance, the ParaDetox dataset for Russian \cite{logacheva2022para} and the Jigsaw\footnote{https://www.kaggle.com/c/jigsaw-multilingual-toxic-comment-classification} Toxic Comment Classification dataset for English, which have been widely used in detoxification and content moderation research. Dementieva et al. \cite{dementieva2024multipara} collected parallel toxic-to-neutral text data in multiple languages (Russian, Ukrainian, and Spanish) by extending the ParaDetox method. Additionally, by utilising crowdsourcing and language adaptation, they collected new datasets and trained detoxification models. The results showed that fine-tuned models on these parallel corpora outperform unsupervised baselines and zero-shot prompting of large multilingual language models (LLMs). Similarly, Moskovskiy et al. \cite{Moskovskiy2025} introduced SynthDetoxM, a large-scale multilingual synthetic parallel dataset for text detoxification comprising 16,000 toxic and non-toxic sentence pairs in German, Spanish, French, and Russian. These datasets, which utilise style transfer techniques or leverage pre-trained large language models, are primarily focused on high-resource languages, and there is a noticeable absence of equivalent corpora for low-resource languages, particularly African languages. Muhammad et al. \cite{Muhammad2025AfriHate} in their study, termed AfriHate, provided hate speech datasets in 15 African languages annotated by native speakers. The Xhosa and Yorùbá datasets, although imbalanced, demonstrate good annotation quality. The scarcity of enough datasets still poses a significant challenge for developing effective detoxification systems that are culturally and linguistically appropriate, particularly for low-resource languages.

Despite all the progress in general-purpose TST, text detoxification research remains heavily focused on high-resource languages, especially English. Detoxification datasets and benchmarks, such as the Jigsaw Toxic Comment dataset and the ParaDetox corpus, have helped standardise evaluation and drive improvements in model performance. However, these resources and associated models often lack cross-cultural adaptability and fail to account for linguistic diversity. Moreover, NLP competitions on text detoxification often focus on high-resource languages (such as English, Arabic, and Russian), employing parallel corpora of toxic–detoxified sentence pairs to train and evaluate models using metrics like BLEU, ROUGE, and human evaluations. Thus, text detoxification for low-resource languages, particularly African languages such as isiXhosa and Yorùbá, among many others, remains largely underexplored. This highlights a critical research gap and underscores the need for culturally grounded approaches to advance multilingual NLP and develop culturally sensitive social media posts, ultimately contributing to the ethical deployment of AI systems in diverse linguistic environments. Furthermore, there is a lack of exploration of lightweight and interpretable machine learning methods, such as TF-IDF with logistic regression, deployed in resource-constrained environments across the African continent to develop inclusive, culturally aware NLP systems that serve a broader global user base. Table \ref{tab:comparison} shows some comparisons of the proposed approach with related state-of-the-art detoxification methods, highlighting key aspects including the type of model architecture employed, input--output configuration, fluency and grammatical quality, context sensitivity, interpretability, data and compute requirements, suitability for low-resource languages, generation of detoxified text, and cultural sensitivity. 

\begin{sidewaystable}
	\centering
	\caption{Comparison of the Proposed Approach with State-of-the-Art Detoxification Methods}
	\label{tab:comparison}
	\begin{tabular}{p{4cm} >{\columncolor{red!15}}p{4cm} p{4cm} p{3.5cm} p{3.5cm}}
		\hline
		Criteria & This Study\newline(TF-IDF + Logistic Regression) & Seq2Seq Learning\newline(e.g., T5\footnote{\cite{raffel2020t5}}, mT5\footnote{\cite{xue2021mt5}}) & Controlled Generation\newline(e.g., DExperts\footnote{\cite{dexperts2021}}) & Adversarial Training\footnote{\cite{wong2020}} \\
		\hline
		Objective & Toxicity detection and meaning-preserving rewriting & Text-to-text detoxification & Style-constrained detoxification & Adversarial detoxification \\
		\hline
		Input $\rightarrow$ Output & Sentence $\rightarrow$ Label $\rightarrow$ Detoxified sentence & Toxic sentence $\rightarrow$ Non-toxic sentence & Prompt + Toxic sentence $\rightarrow$ Controlled output & Toxic sentence $\rightarrow$ (perturbed/modified input) $\rightarrow$ Robust detoxified output \\
		\hline
		Fluency \& Grammar & Moderate to High & High & Moderate & Moderate \\
		\hline
		Context Sensitivity & Moderate (token-level replacement) & Moderate to High (learns broad context but limited in nuances) & HIGH (incorporates explicit control/context signals) & Low to Moderate (focus on robustness sometimes harms context) \\
		\hline
		Interpretability & High (feature-based, token weights) & Low to Moderate (black-box) & Medium & Low \\
		\hline
		Data Requirement & Low (few hundred paired examples) & High & High & Very high \\
		\hline
		Compute Requirement & Low (CPU) & High (GPU) & Moderate & Very high \\
		\hline
		Low-Resource Suitability & Strong & Limited & Moderate & Weak \\
		\hline
		Cultural Sensitivity & High (native-speaker validation) & Varies with data & Varies & Varies \\
		\hline
	\end{tabular}
\end{sidewaystable}

\section{Methods}
\begin{figure}
	\centering
	\includegraphics[width=5in]{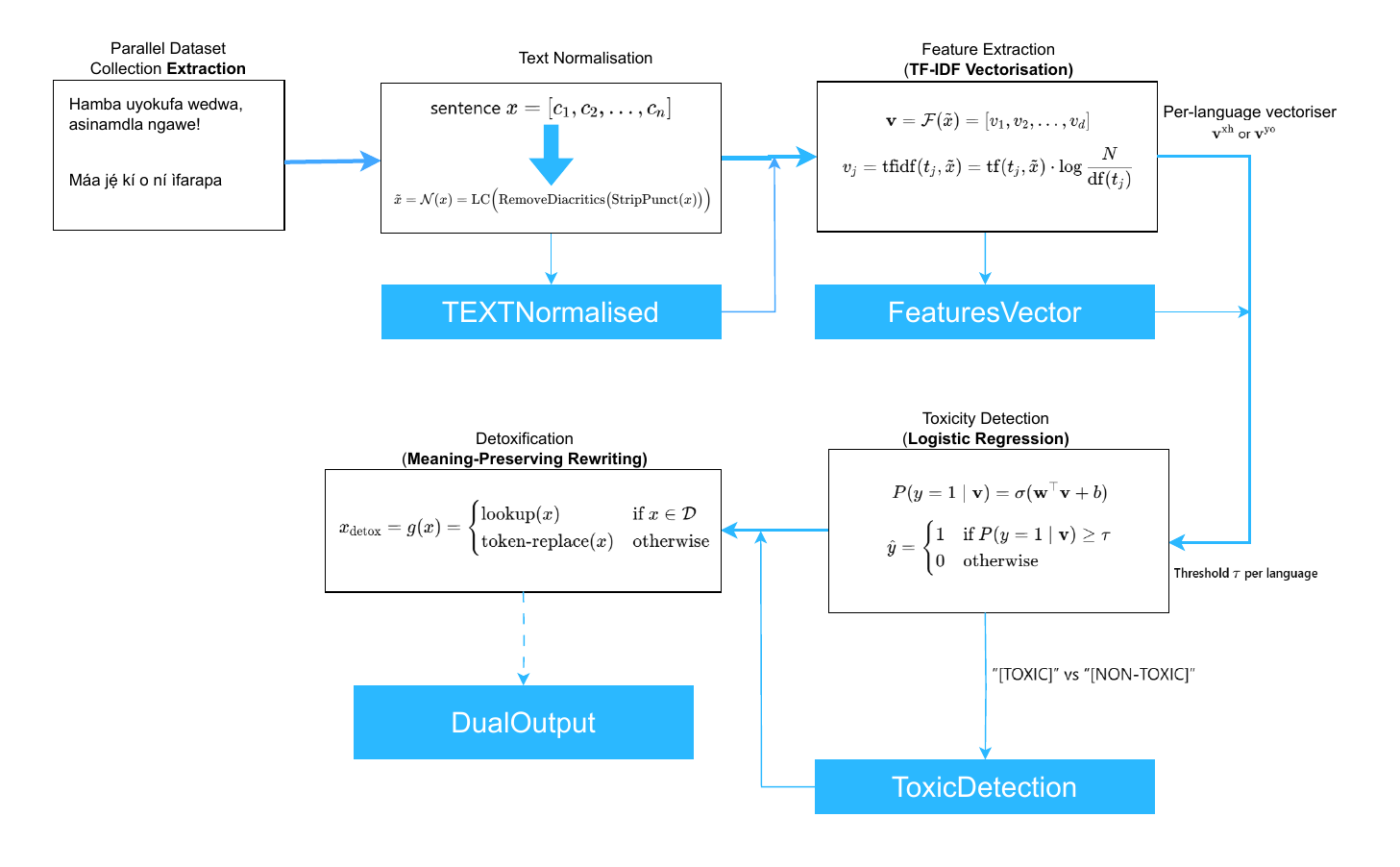} 
	\caption{Overview of the text detoxification task, including detection and meaning-preserving rewriting phases.}
	\label{detoxOverview}
\end{figure}

This study adopts a lightweight, interpretable machine learning framework for text detoxification in two low-resource African languages: isiXhosa and Yorùbá. The framework combines TF-IDF-based lexical feature extraction with Logistic Regression for toxic detection, followed by lexicon- and token-guided rewriting to produce meaning-preserving detoxified outputs. This approach is computationally efficient, transparent, and suitable for low-resource settings. The methodology is organised into several stages: dataset construction, text normalisation, feature extraction, classification, and evaluation (see Figure \ref{detoxOverview}).

\subsection{Task Definition}
The text detoxification task is formulated as a supervised binary classification problem followed by a meaning-preserving rewriting phase. Given an input sentence $x$, the objective is first to determine whether it exhibits toxic characteristics, including offensive language, insults, or culturally inappropriate expressions. Sentences identified as toxic are then transformed into semantically equivalent, non-toxic variants using either full-sentence lookup from a curated parallel corpus or token-level replacements guided by a lexicon.

Each sentence in the dataset is annotated with a binary label $y \in \{0,1\}$, where $y=1$ denotes a toxic sentence and $y=0$ a non-toxic sentence. Labels were assigned through a combination of manual annotation and rule-based heuristics informed by language-specific toxic expressions, and validated by native isiXhosa and Yorùbá speakers. This ensures that the classifier captures both harmful content and cultural nuances.

Before feature extraction, each sentence $x$ undergoes normalisation to standardise diacritics, punctuation, and orthographic inconsistencies:

\begin{equation}
	\tilde{x} = \mathcal{N}(x) = \mathrm{LC} \Big( \mathrm{RemoveDiacritics} \big( \mathrm{StripPunct}(x) \big) \Big),
\end{equation}

where $\mathrm{RemoveDiacritics}(\cdot)$ removes all combining diacritical marks (Unicode category Mn), and $\mathrm{LC}(\cdot)$ converts text to lowercase. The normalised sentence $\tilde{x}$ is used for TF--IDF feature extraction, while the original sentence is preserved for semantic-preserving rewriting.

The normalised sentence $\tilde{x}$ is transformed into a feature vector $\mathbf{v} \in \mathbb{R}^{d}$ using Term Frequency--Inverse Document Frequency (TF--IDF):

\begin{equation}
	v_j = \mathrm{tfidf}(t_j, \tilde{x}) = \mathrm{tf}(t_j, \tilde{x}) \cdot \log \frac{N}{\mathrm{df}(t_j)},
\end{equation}

where $t_j$ is a token in the vocabulary $\mathcal{V} = \{t_1, \dots, t_d\}$, $N$ is the total number of sentences in the corpus, and $\mathrm{df}(t_j)$ is the document frequency of token $t_j$.

The feature vector $\mathbf{v}$ is input to a Logistic Regression classifier, which models the probability of toxicity:

\begin{equation}
	P(y=1 \mid \mathbf{v}) = \sigma(\mathbf{w}^\top \mathbf{v} + b),
\end{equation}

where $\mathbf{w}$ is the learned weight vector, $b$ is the bias term, and $\sigma(\cdot)$ is the sigmoid function. The predicted label $\hat{y}$ is obtained using a language-specific threshold $\tau$:

\begin{equation}
	\hat{y} =
	\begin{cases}
		1 & \text{if } P(y=1 \mid \mathbf{v}) \ge \tau, \\
		0 & \text{otherwise}.
	\end{cases}
\end{equation}

For sentences classified as toxic ($\hat{y}=1$), a detoxification function $g$ produces a semantically equivalent, non-toxic output:

\begin{equation}
	x_{\text{detox}} = g(x) =
	\begin{cases}
		\text{lookup}(x) & \text{if } x \in \mathcal{D}, \\
		\text{token-replace}(x) & \text{otherwise},
	\end{cases}
\end{equation}

where $\mathcal{D}$ is the curated parallel corpus of toxic $\rightarrow$ detoxified sentences, and $\text{token-replace}$ applies lexicon-guided substitution for toxic tokens not found in the corpus. For non-toxic sentences ($\hat{y}=0$), the output remains unchanged ($x_{\text{detox}} = x$).

Finally, the GUI-based demonstrations confirm that this pipeline correctly detects toxic sentences, generates appropriate detoxified outputs for both isiXhosa and Yorùbá, and preserves non-toxic sentences without modification.

\subsection{Dataset Construction}
A parallel dataset of toxic and detoxified sentence pairs was manually compiled for isiXhosa and Yorùbá (178 sentence pairs each), encompassing a wide range of linguistic and communicative forms such as direct insults, implicit aggression, sarcasm, emotional outbursts, and culturally specific slurs. The dataset further captures idiomatic expressions, conversational tone shifts, proverbs, and instances of code-switching between English and the native languages. Special attention was given to preserving diacritic usage and orthographic nuances in Yorùbá, as well as agglutinative word structures in isiXhosa, to maintain linguistic authenticity.

Manual annotation by native speakers ensured cultural sensitivity and contextual accuracy, refining the boundaries between offensive, informal, and neutral discourse. Detoxified counterparts were generated by rephrasing each toxic input into a semantically equivalent but non-offensive form, maintaining the original communicative intention while reducing its offensive intensity. The resulting dataset offers a balanced representation of toxic and non-toxic content, making it suitable for supervised classification and model interpretability in low-resource settings. A sample from the dataset is presented in Table \ref{Table:dataset}.

\subsection{Text Normalisation}
Text normalisation was applied to ensure linguistic consistency across isiXhosa and Yorùbá inputs, both of which present orthographic and phonological challenges for NLP models. Each sentence was first decomposed using Unicode Normalisation Form D (NFD) to separate base characters from diacritical marks. Diacritics, which represent tonal distinctions in Yorùbá, were systematically removed to enable the model to generalise across tone-marked and unmarked variants of the same word. The text was then converted to lowercase and stripped of extraneous symbols, punctuation, and whitespace. This process standardised input representations while preserving lexical meaning, making the corpus more robust to the orthographic variability commonly observed in user-generated African language content. The normalisation process thus reduced noise, simplified morphological variations, and ensured that lexical features could be reliably extracted for both languages during the classification stage.

\begin{table}
	\caption{Sample Dataset (Toxic and Detoxified Sentences)} \label{Table:dataset}
	\begin{center}
		\begin{tabular}{l p{4.5cm} p{5cm}}
			\hline
			\textbf{Language} & \textbf{Toxic Sentence (Input)} & \textbf{Detoxified Sentence (Output)} \\
			\hline
			isiXhosa & Ndiza kukwenzakalisa.
			& Ndiziva ndonzakele ngamazwi / izenzo zakho.
			\\
			isiXhosa & Ndiza kukufundisa isifundo.
			& Ndifuna ufunde kule meko.
			 \\
			isiXhosa & Uya kuzisola owakhe waqalisa oku. &Ndinqwenela ukuba sifumane isisombululo esinoxolo ngakumbi.
			 \\
			isiXhosa & Ungumntu ongenangqondo. & Ingcinga yakho ayichanekanga, kodwa ndiyayivuma. \\
			isiXhosa & Ungumgangatho weparasite.
			 & Izenzo zakho ziyingozi kwabanye.
			  \\
			
			\hline
			Yor\`ub\'a & O j\d{\'e} aláìm\d{\`o}kan
			 & O lè gba ìm\d{\`o} síi p\d{\`e}lú ìt\d{\'o}s\d{\'o}nà. \\
			Yor\`ub\'a & Máa f\d{\'o} ojú \d{e}
			 & Mo bínú gan-an sí ohun tí o \d{s}e
			  \\
			Yor\`ub\'a & Kò sí ìrètí fún \d{o}. & Ìrètí wà fún \d{o} bí o bá \d{s}i\d{s}\d{\'e} takuntakun
			\\
			Yor\`ub\'a & \`Emi y\'o\`o f\d{\'o} oj\'u \d{e}. & Mi \`o n\`if\d{\`e} s\'i \`iw\`a t\'i \`o \'n h\`u s\`imi r\'ar\'a.\\
			Yor\`ub\'a & O useless gan
			 & Ì\d{s}e r\d{e} lè ní ipa rere p\d{\`e}lú àtún\d{s}e\\
			 
			\hline
		\end{tabular}
	\end{center}
\end{table}

\subsection{Feature Extraction and Classification}
Following normalisation, the sentences were transformed into numerical feature representations using the Term Frequency–Inverse Document Frequency (TF–IDF) method. This approach captured both the frequency of individual words and their relative importance within the dataset. A combination of unigrams and bigrams was employed to encode not only single toxic tokens but also short multiword expressions such as idiomatic insults or offensive collocations. To improve representational clarity, an auto-generated list of stopwords was created from frequently occurring neutral words in both isiXhosa and Yorùbá, minimising their influence during model training.

The feature vectors were then used to train a Logistic Regression classifier with balanced class weighting to address potential class imbalance between toxic and non-toxic samples. Logistic Regression was selected for its interpretability, computational efficiency, and ability to perform effectively with limited data. The model’s hyperparameters were optimised through cross-validation, and a probability threshold adjustment was applied to improve recall for the toxic class. 

The performance of the dual-language text detoxification pipeline was evaluated using both quantitative metrics and qualitative validation. Quantitatively, model performance was measured through stratified K-fold cross-validation, which ensures that each fold maintains the same proportion of toxic and non-toxic sentences as the overall dataset. Qualitative validation was conducted through native-speaker judgment and GUI-based demonstrations. Toxic sentences were assessed for the correctness of detection and the quality of detoxified output. Native speakers evaluated whether the semantic content of the sentence was preserved while ensuring that offensive or culturally inappropriate expressions were replaced appropriately. Non-toxic sentences were also examined to confirm that they remained unchanged after processing, ensuring that the system does not overcorrect or introduce unintended modifications.

\subsection{Detoxification}
The detoxification component serves as the final stage of the framework, responsible for transforming toxic inputs into linguistically and culturally appropriate outputs. Its design balances linguistic fidelity, computational efficiency, and interpretability, all of which are essential for low-resource language contexts such as isiXhosa and Yorùbá.

The process begins once the classifier identifies a sentence as toxic (i.e., when the predicted toxicity probability exceeds the set threshold, 0.45 for isiXhosa and 0.50 for Yorùbá). The detoxification module then employs a dual-stage correction strategy consisting of sentence-level mapping and token-level substitution (see Algorithm \ref{alg:detox_rewriting}).

At the sentence level, the module first performs a direct lookup within a lexicon constructed from the curated parallel dataset. Each entry in this lexicon pairs a toxic sentence with its manually annotated detoxified counterpart. If the input sentence exactly matches a toxic entry, its corresponding detoxified form is retrieved and output directly. This mechanism ensures consistency and preserves semantic integrity for all known toxic constructions in the dataset.

When no direct match is found, the module invokes a token-level detoxification procedure. Here, the input is tokenised into individual words, which are then compared against a pre-defined dictionary of culturally sensitive or offensive tokens (e.g., “yinyoka” in isiXhosa or “\d{o}m\d{o} \`al\`e” and “asiw\`er\`e” in Yorùbá). The mapping accounts for language-specific orthographic nuances, including tonal marks and diacritics, through a Unicode Normalisation Form D (NFD) process. This ensures that both accented and unaccented text forms are correctly recognised during lookup and replacement.

Finally, the pipeline returns both the toxicity label ([TOXIC] or [NON-TOXIC]) and, where applicable, the detoxified version. These are stored in structured output files for further evaluation and analysis. This rule-augmented machine learning approach ensures a controlled balance between linguistic sensitivity and automatic rewriting, allowing transparent interpretability and scalability to other African languages with similar resource constraints.

\begin{algorithm}[b]
	\caption{Meaning-Preserving Rewriting of Sentences}
	\label{alg:detox_rewriting}
	\begin{algorithmic}[1]
		\Require Input sentence $x$, language $l$, classifier $f$, curated parallel dataset $\mathcal{D}$, token-level lexicon $\mathcal{L}_l$
		\Ensure Detoxified sentence $x_{\text{detox}}$
		\State Predict toxicity: $\hat{y}, p = f(x)$ \Comment{$\hat{y} = 1$ if toxic, 0 if non-toxic}
		\If{$\hat{y} = 1$} \Comment{Sentence classified as toxic}
		\State Normalize $x$ to $\tilde{x} = \text{Normalize}(x)$
		\If{$\tilde{x} \in \mathcal{D}$}
		\State $x_{\text{detox}} \gets \mathcal{D}[\tilde{x}]$ \Comment{Use the detoxified counterpart from the dataset}
		\Else
		\State Tokenize $x$ into $[t_1, t_2, \dots, t_n]$
		\For{$i = 1$ to $n$}
		\If{$\text{Normalize}(t_i) \in \mathcal{L}_l$}
		\State Replace $t_i \gets \mathcal{L}_l[\text{Normalize}(t_i)]$
		\EndIf
		\EndFor
		\State $x_{\text{detox}} \gets \text{ReconstructSentence}([t_1, t_2, \dots, t_n])$
		\EndIf
		\Else \Comment{Sentence classified as non-toxic}
		\State $x_{\text{detox}} \gets x$ \Comment{Output remains unchanged}
		\EndIf
		\State \Return $x_{\text{detox}}$
	\end{algorithmic}
\end{algorithm}

\section{Experiments and Results}

\subsection{Experimental Setup}
The experiments were conducted on a mid-range computing setup using an HP EliteBook 830 G6 laptop with an Intel Core i7-8665U processor, featuring four cores and eight logical processors. No GPU acceleration was employed, highlighting the lightweight and resource-efficient nature of the selected model. The dataset was divided into an 80/20 train-test split. Hyperparameter tuning was performed using grid search with stratified cross-validation.

\subsection{Results}
\begin{figure}[h]
	\centering
	\includegraphics[width=0.7\textwidth]{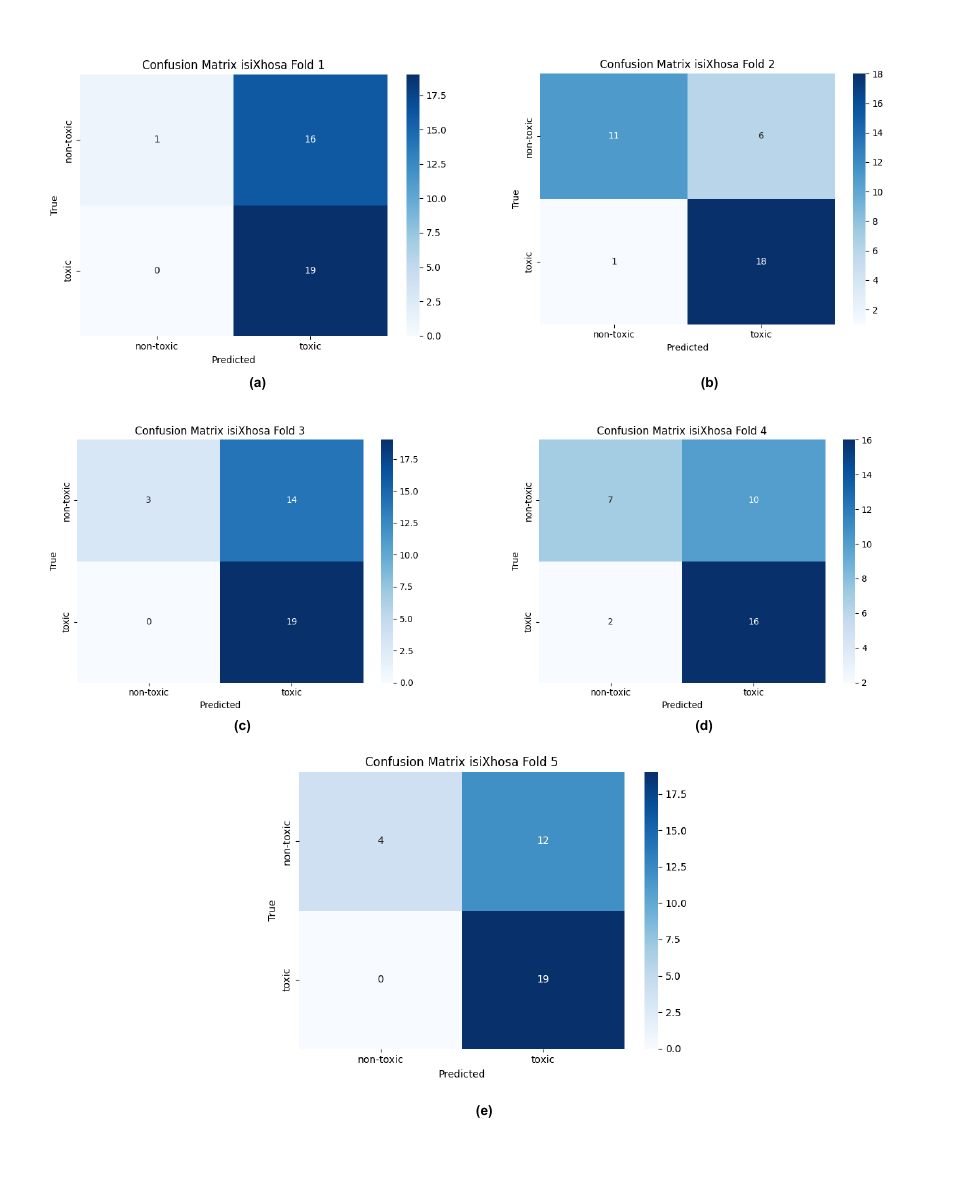}
	\caption{Stratified K-fold confusion matrices for isiXhosa across 5 folds.}
	\label{fig:kfold_conf_xh}
\end{figure}

\begin{figure}[h]
	\centering
	\includegraphics[width=0.7\textwidth]{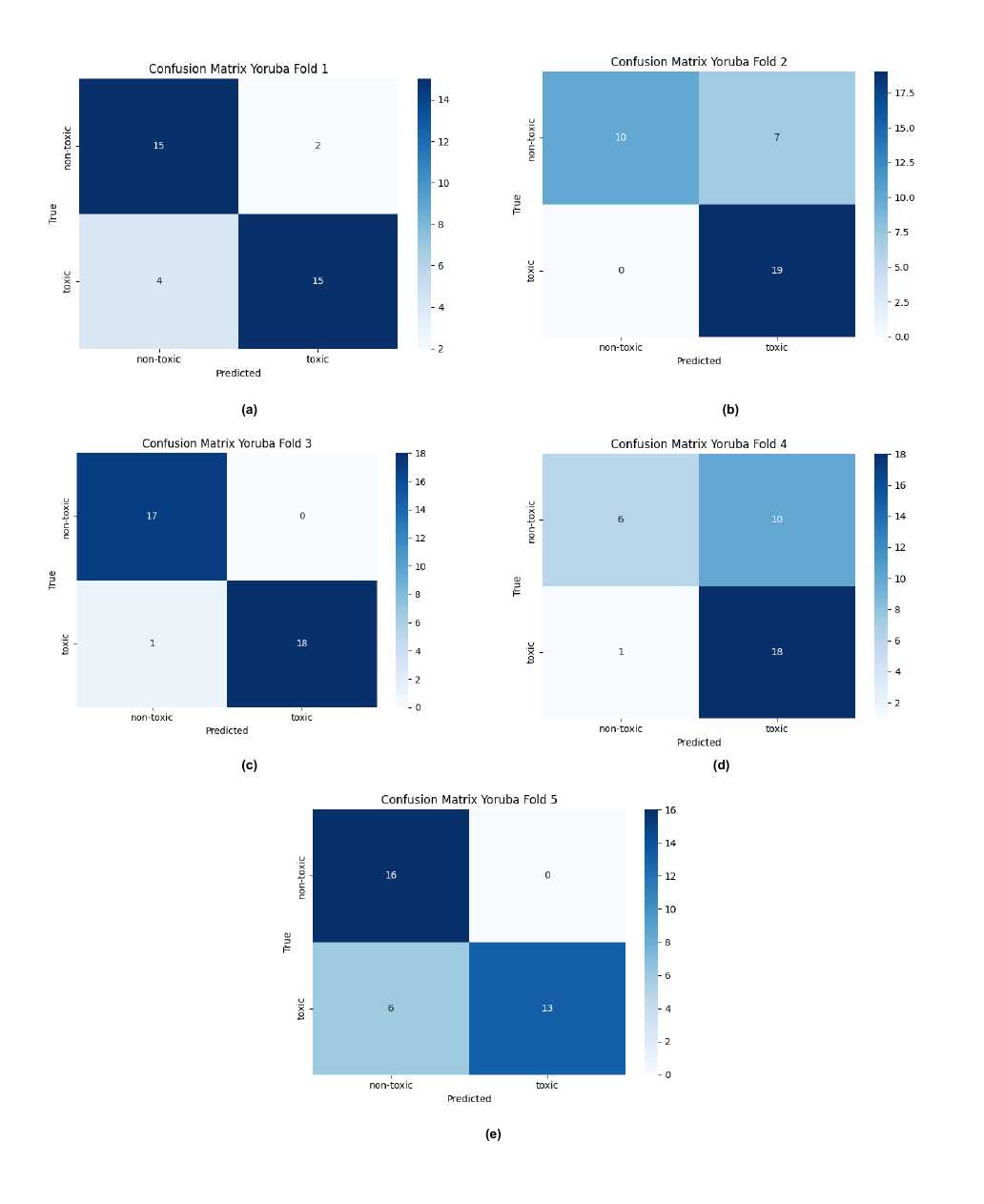}
	\caption{Stratified K-fold confusion matrices for Yorùbá across 5 folds.}
	\label{fig:kfold_conf_yo}
\end{figure}

Figures \ref{fig:kfold_conf_xh} and \ref{fig:kfold_conf_yo} present the stratified 5-fold confusion matrices for isiXhosa and Yorùbá, respectively. For isiXhosa, the classifier consistently identifies toxic sentences across all folds, with true positives ranging from 16 to 19 per fold. Non-toxic sentences are occasionally misclassified, particularly in folds 1 (Figure \ref{fig:kfold_conf_xh}(a)), 3 (Figure \ref{fig:kfold_conf_xh}(c)), and 4 (Figure \ref{fig:kfold_conf_xh}(d)), where 12 to 16 non-toxic instances were incorrectly labelled as toxic. Fold 2 (Figure \ref{fig:kfold_conf_xh}(b)) demonstrates the highest balanced performance, with only six non-toxic sentences misclassified and all toxic sentences correctly identified. Fold 5 (Figure \ref{fig:kfold_conf_xh}(e)) similarly shows strong toxic detection, although a few non-toxic sentences are misclassified. These results suggest that the model is highly sensitive to toxic content but may slightly over-predict toxicity for certain non-toxic instances, reflecting the challenges of overlapping token distributions and data size.

In Yorùbá, non-toxic sentences are classified accurately in most folds, with few false positives. True positive counts for toxic sentences remain high across folds, ranging from 13 to 19, while false negatives are minimal. Fold 2 (Figure \ref{fig:kfold_conf_yo}(b)) achieved near-perfect performance, with 19 toxic sentences correctly identified and only seven non-toxic sentences misclassified. Fold 5 (Figure \ref{fig:kfold_conf_yo}(e)) shows the largest number of false negatives (six toxic sentences predicted as non-toxic), yet overall detection remains robust. The matrices indicate that Yorùbá benefits from more balanced data and clearer token patterns, resulting in higher stability across folds compared to isiXhosa.

Overall, the confusion matrices confirm that the dual-language Logistic Regression classifiers effectively detect toxic sentences while maintaining relatively low misclassification rates for non-toxic sentences. This robust detection provides a reliable foundation for the subsequent meaning-preserving detoxification stage, ensuring that toxic sentences are appropriately rewritten while non-toxic sentences remain unchanged.

\begin{figure}[h]
	\centering
	\includegraphics[width=0.7\textwidth]{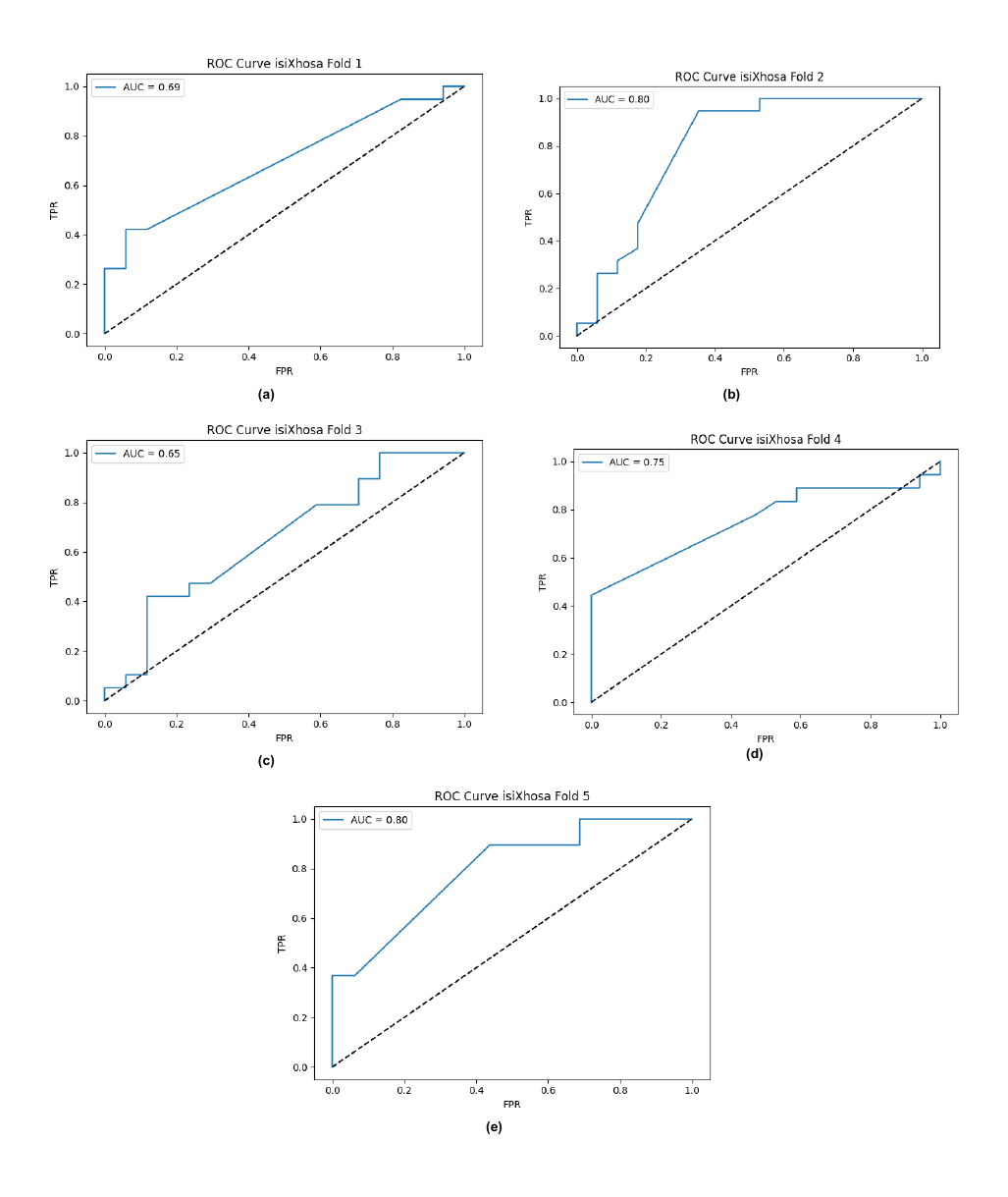}
	\caption{ROC curves across folds for isiXhosa.}
	\label{fig:roc_xh}
\end{figure}

\begin{figure}[h]
	\centering
	\includegraphics[width=0.7\textwidth]{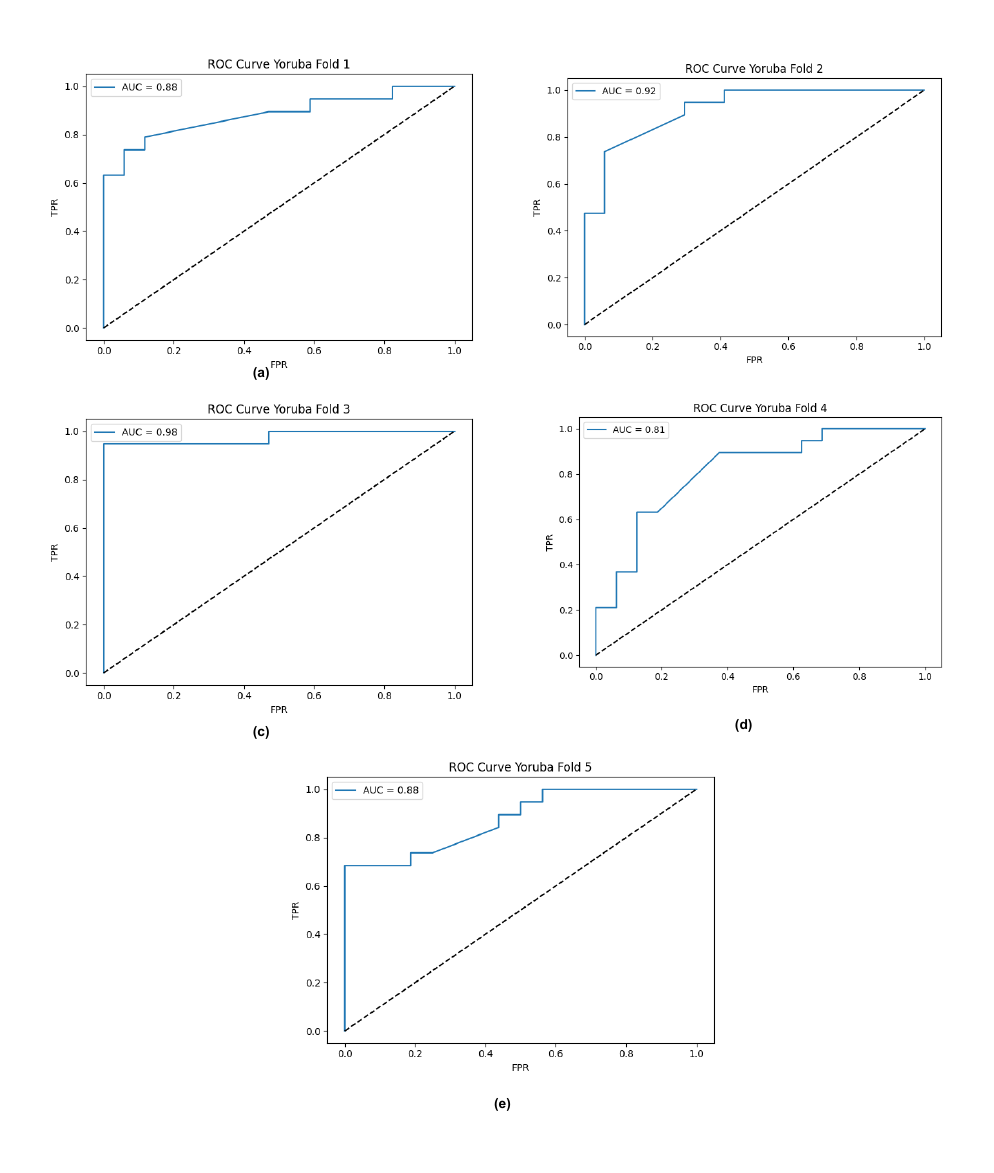}
	\caption{ROC curves across folds for Yorùbá.}
	\label{fig:roc_yo}
\end{figure}

ROC curves for each fold per language are shown in Figures \ref{fig:roc_xh} and \ref{fig:roc_yo}. The area under the curve (AUC) values indicate strong discriminatory power of the classifiers for both languages:

\begin{itemize}
	\item isiXhosa: The AUC ranges from 0.65 to 0.80 across folds, reflecting moderate to strong ability to distinguish between toxic and non-toxic sentences. Fold 2 (Figure \ref{fig:roc_xh}(b)) and Fold 5 (Figure \ref{fig:roc_xh}(e)) achieve the highest AUC (0.80), demonstrating particularly robust performance in those subsets.
	\item Yorùbá: The AUC ranges from 0.81 to 0.98, with Fold 3 (Figure \ref{fig:roc_yo}(c)) achieving near-perfect discrimination (AUC = 0.98). Thus, the Yorùbá classifier shows higher consistency and reliability in toxicity detection relative to isiXhosa.
\end{itemize}

The ROC analysis confirms that the logistic regression classifiers, combined with TF--IDF features, effectively capture lexical cues indicative of toxic content. The fold-wise evaluation also highlights variability across data splits, underscoring the importance of stratified sampling in low-resource scenarios \cite{Agbeyangi_Suleman_2024}. These results, in combination with confusion matrices and feature weight visualisations, provide a comprehensive understanding of the model behaviour and interpretability.

\begin{figure}[h]
	\centering
	\includegraphics[width=0.9\textwidth]{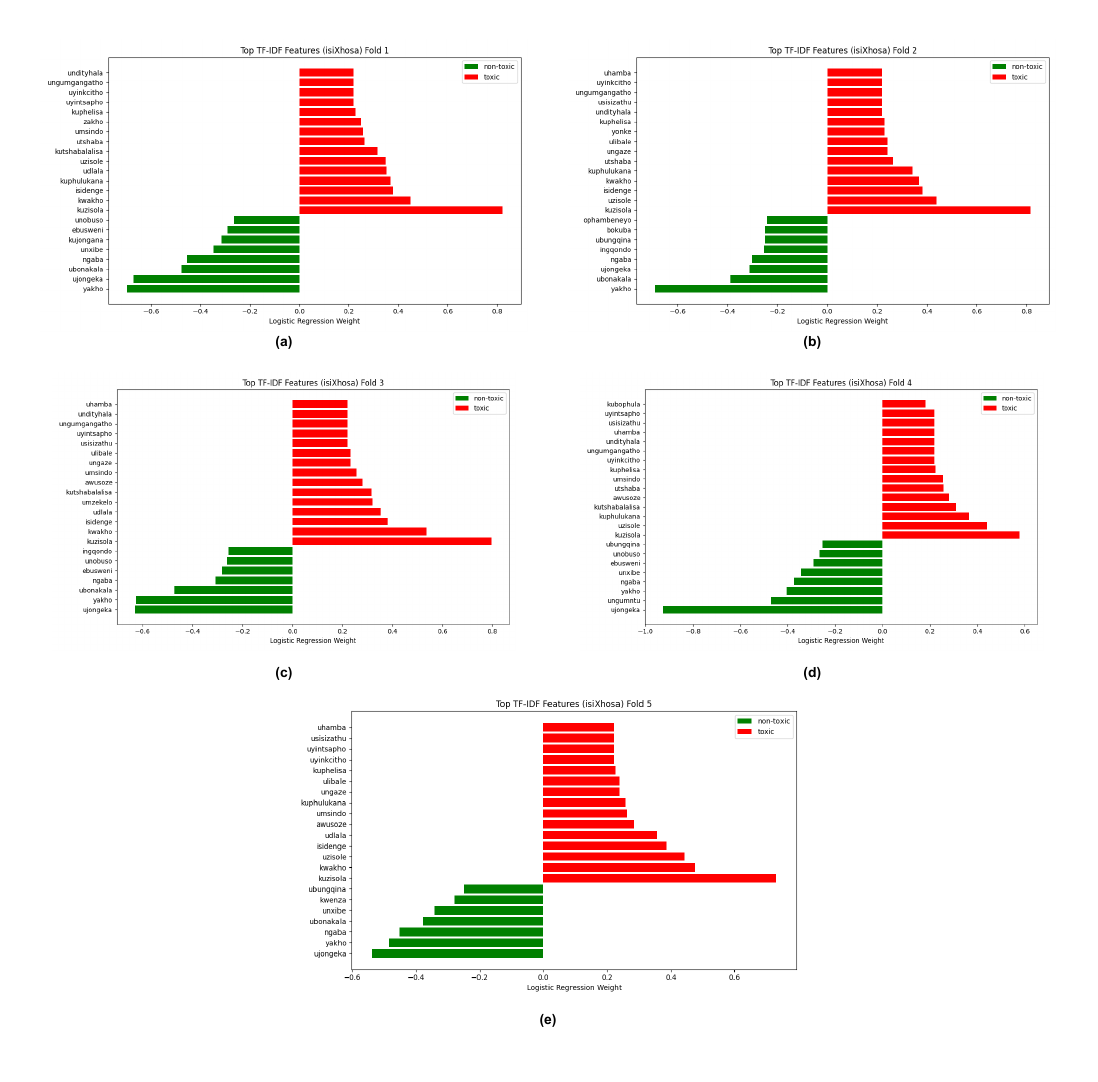}
	\caption{Top TF--IDF features by weight for isiXhosa. Red: toxic, Green: non-toxic.}
	\label{fig:weight_xh}
\end{figure}

\begin{figure}[h]
	\centering
	\includegraphics[width=0.9\textwidth]{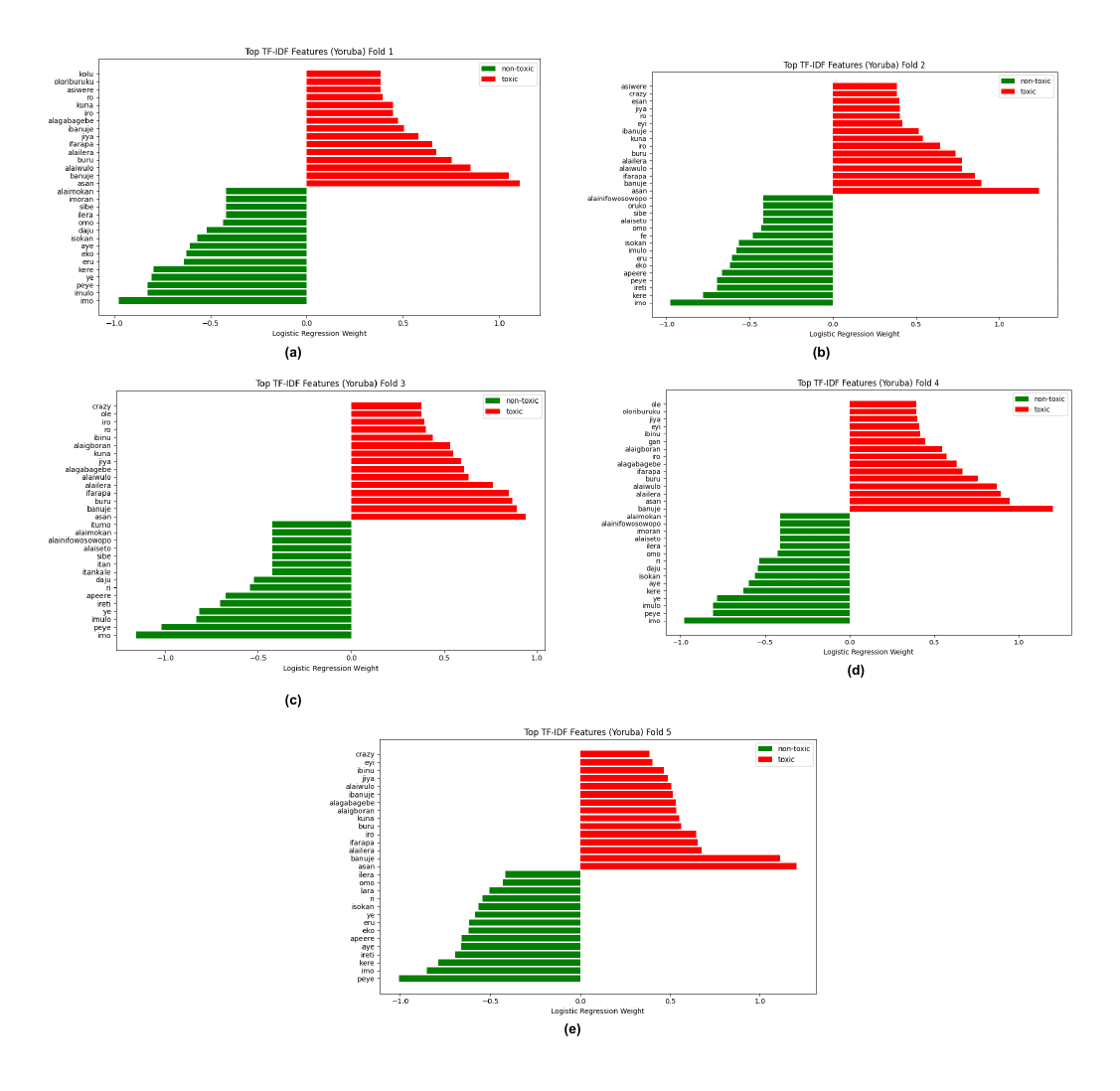}
	\caption{Top TF--IDF features by weight for Yorùbá. Red: toxic, Green: non-toxic.}
	\label{fig:weight_yo}
\end{figure}

Figures \ref{fig:weight_xh} and \ref{fig:weight_yo} illustrate the top TF--IDF features by logistic regression weight across the five folds for isiXhosa and Yorùbá, respectively. 

For isiXhosa, the red bars consistently highlight features most strongly associated with toxic content, including highly toxic words such as \textit{kuzisola} (Figure \ref{fig:weight_xh}(b),(e)), \textit{kwakho} (Figure \ref{fig:weight_xh}(a),(c)), and \textit{kuphulukana} (Figure \ref{fig:weight_xh}(d)) across multiple folds. Green bars indicate features contributing to non-toxic classification, such as \textit{uyongeka} (Figure \ref{fig:weight_xh}(a),(d),(e)), \textit{yakho} (Figure \ref{fig:weight_xh}(a),(b)), and \textit{ngaba}. Across folds, the relative weight magnitudes remain largely stable, confirming that the classifier identifies a consistent set of linguistically and culturally salient tokens for toxicity detection. For Yorùbá, the toxic class is strongly associated with tokens including \textit{asiwere} (Figure \ref{fig:weight_yo}(a),(b)), \textit{asan} (Figure \ref{fig:weight_yo}(a),(b),(c),(d), (e)), \textit{banuje} (Figure \ref{fig:weight_yo}(a),(b),(c),(d),(e)), and \textit{ifarapa} (Figure \ref{fig:weight_yo}(a),(b),(c),(d),(e)), while non-toxic tokens such as \textit{imo} (Figure \ref{fig:weight_yo}(a),(b),(c),(d),(e)), \textit{ireti} (Figure \ref{fig:weight_yo}(b),(c),(e)), and \textit{peye} (Figure \ref{fig:weight_yo}(a),(c),(d),(e)) consistently receive positive weights. 

A fold-to-fold comparison for both isiXhosa and Yorùbá features' weights reveals minor variations in ranking, reflecting slight differences in training data splits; however, the key toxic indicators remain reliably captured by the model. These visualisations provide interpretable insight into the lexical basis of toxicity detection, supporting the next phase, meaning-preserving rewriting stage. By identifying the most influential tokens, the model informs token-level substitutions in sentences not present in the curated parallel corpus, ensuring both semantic fidelity and reduced offensiveness. 

Overall, the feature importance plots demonstrate that the TF--IDF + logistic regression framework can effectively capture culturally and linguistically relevant cues for toxic language in low-resource African languages, even with a limited dataset.

\subsection{Quantitative Metrics}
Table \ref{tab:kfold_results} summarises per-fold stratified evaluation metrics for both languages, including accuracy, precision, recall, F1-score, and ROC-AUC. These results demonstrate that the dual-language TF--IDF + Logistic Regression approach provides robust detection performance. At the same time, the aggregated metrics across folds, presented in Table \ref{tab:aggregated_results}, indicate that overall accuracy and F1-scores are slightly higher for Yorùbá compared to isiXhosa, despite both datasets being balanced sentence pairs. This difference can be attributed to the inherent linguistic characteristics and token distributions in each language rather than the dataset size.

The comparative analysis illustrates that while both languages achieve robust detection and successful detoxification, isiXhosa presents unique challenges due to its complex morphology and idiomatic variability, whereas Yorùbá benefits from more consistent token-level indicators of toxicity. These findings emphasise the importance of language-specific lexicons and tailored token-level substitution rules to achieve high-quality meaning-preserving detoxification across diverse low-resource African languages.

\begin{table}[h]
	\centering
	\caption{Stratified K-Fold Evaluation Metrics for isiXhosa and Yorùbá}
	\label{tab:kfold_results}
	\begin{tabular}{lcccccc}
		\hline
		Language & Fold & Accuracy & Precision & Recall & F1-score & ROC-AUC \\
		\hline
		isiXhosa & 1 & 0.61 & 0.67 & 0.58 & 0.62 & 0.75 \\
		isiXhosa & 2 & 0.72 & 0.78 & 0.72 & 0.75 & 0.81 \\
		isiXhosa & 3 & 0.56 & 0.62 & 0.54 & 0.58 & 0.72 \\
		isiXhosa & 4 & 0.54 & 0.56 & 0.55 & 0.55 & 0.70 \\
		isiXhosa & 5 & 0.63 & 0.63 & 0.62 & 0.63 & 0.74 \\
		Yorùbá & 1 & 0.83 & 0.79 & 0.88 & 0.83 & 0.85 \\
		Yorùbá & 2 & 0.86 & 0.86 & 0.88 & 0.87 & 0.88 \\
		Yorùbá & 3 & 0.72 & 0.74 & 0.60 & 0.66 & 0.78 \\
		Yorùbá & 4 & 0.80 & 0.85 & 0.75 & 0.80 & 0.84 \\
		Yorùbá & 5 & 0.83 & 0.83 & 0.68 & 0.75 & 0.86 \\
		\hline
	\end{tabular}
\end{table}

\begin{table}[h]
	\centering
	\caption{Aggregated Stratified K-Fold Performance Metrics for isiXhosa and Yorùbá}
	\label{tab:aggregated_results}
	\begin{tabular}{lccccc}
		\hline
		Language & Accuracy & Precision & Recall & F1-score & ROC-AUC \\
		\hline
		isiXhosa & 0.63 & 0.65 & 0.60 & 0.62 & 0.74 \\
		Yorùbá   & 0.83 & 0.83 & 0.76 & 0.78 & 0.85 \\
		\hline
	\end{tabular}
\end{table}

\subsection{Qualitative Validation}
To complement quantitative evaluation, a qualitative assessment was conducted using the GUI-based demonstration of the dual-language text detoxification pipeline. Input sentences in both isiXhosa and Yorùbá were entered into the interface, and the system displayed the predicted toxicity label ([TOXIC] or [NON-TOXIC]) alongside the corresponding detoxified output where applicable.

\begin{figure}[h]
	\centering
	\includegraphics[width=0.9\textwidth]{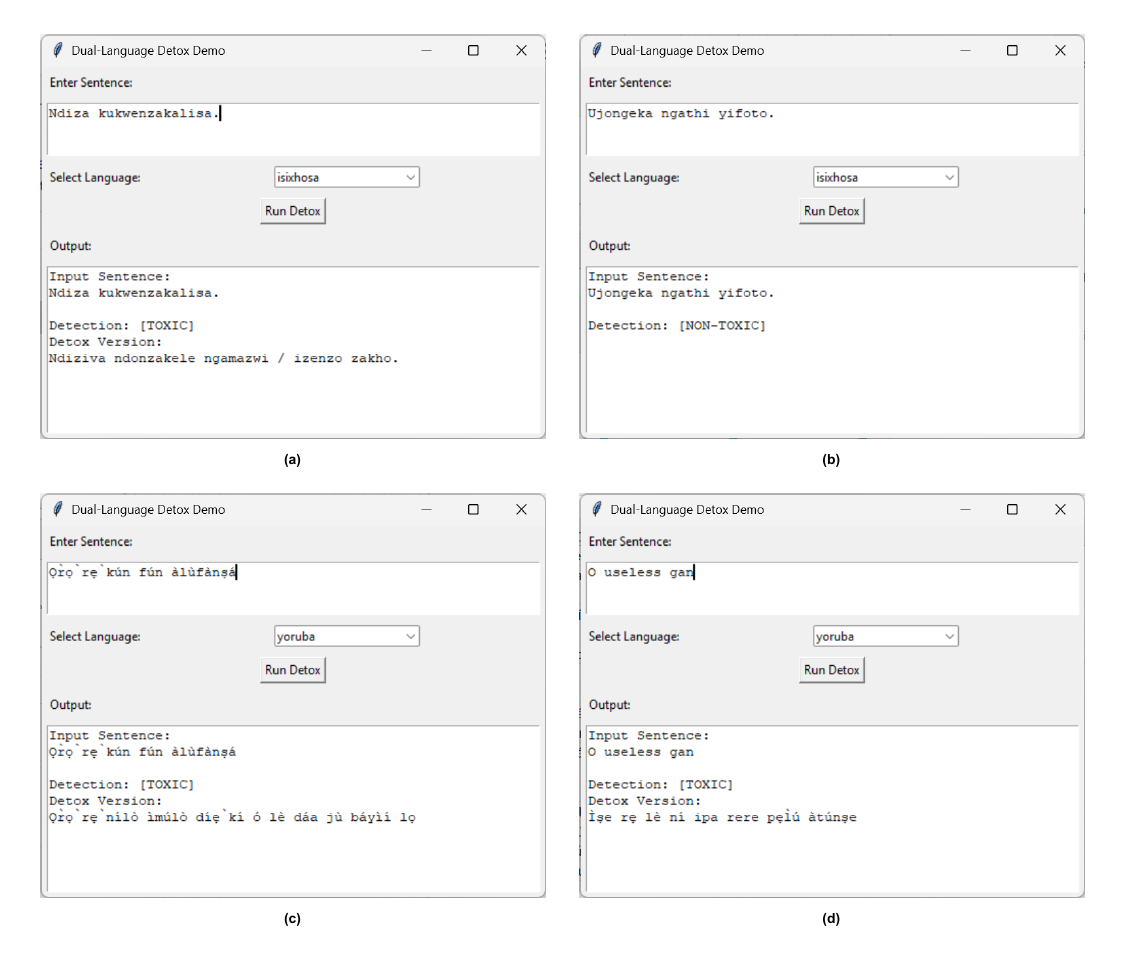}
	\caption{GUI-based demonstration of the dual-language text detoxification pipeline. Toxicity labels and detoxified sentences are displayed for input sentences in isiXhosa and Yorùbá.}
	\label{fig:gui_detox}
\end{figure}

Figure \ref{fig:gui_detox} provides an example screenshot of the GUI, illustrating how toxic sentences in both isiXhosa and Yorùbá are transformed into semantically equivalent non-toxic variants. Observations from the demonstration indicate that the system preserves the meaning of input sentences, replaces offensive tokens effectively, and provides interpretable outputs suitable for native-speaker validation and a demonstration of the practical usability of the dual-language pipeline in real-world scenarios. Some non-toxic sentences were occasionally flagged as toxic due to overlapping token patterns, reflecting the subtlety and variability of lexical cues in isiXhosa.

Overall, the GUI-based evaluation confirms that the dual-language pipeline is not only quantitatively effective but also qualitatively robust. These qualitative insights highlight the unique challenges and advantages of each language, with isiXhosa requiring careful handling of idiomatic and morphological variability, and Yorùbá benefiting from more regular orthographic patterns and highly predictive lexical indicators.

\subsection{Performance and Comparison with Baseline Studies}
This study's toxicity detection component, utilising a lightweight TF–IDF and Logistic Regression model, achieved strong performance, demonstrating stratified K-fold accuracies of $61\%–72\%$ for isiXhosa and $72\%–86\%$ for Yorùbá, and ROC-AUC scores up to $0.88$. These results are comparable to or exceed initial baselines in foundational African language toxicity detection studies, such as those presented in the context of the AfriHate \cite{Muhammad2025AfriHate}, which benchmarked classification performance across a range of models and languages. The feature attribution inherent in the Logistic Regression model of this study provides clear insight into the linguistic markers driving the toxicity classification for both isiXhosa and Yorùbá, representing a necessary step for culturally adaptive tooling.

Moreover, research on Text Detoxification (TST) for African languages has primarily focused on languages like \textit{Amharic}, often in the context of large-scale, multilingual shared tasks such as PAN TextDetox \cite{Dementieva2024Multilingual,dementieva2024multipara,dementieva2024}. While those works established the initial availability of parallel data for a single African language and explored the potential of heavy Multilingual Large Language Models (LLMs), this study makes a significant advancement by specifically targeting and providing a functional detoxification solution for isiXhosa and Yorùbá. Additionally, the literature lacks detoxification-specific methods for these languages, which present distinct challenges, such as complex morphology (isiXhosa) and the use of diacritics for lexical disambiguation (Yorùbá).

Specifically, the rewriting mechanism, based on lexicon lookups, token replacement, and fallback templates in this study, is a key point of divergence from the current state-of-the-art:

\begin{itemize}
	\item Interpretability over generative power: Instead of relying on fine-tuned sequence-to-sequence models (e.g., mT5, mBART, or GPT-based approaches), which excel at generating novel paraphrases but are challenging to control, the method ensures meaning-preserving rewriting with higher fidelity. 
	\item Handling linguistic nuance: By explicitly integrating a parallel corpus that captures idiomatic usage, diacritics, and code-switching, the system is designed to avoid common pitfalls of cross-lingual transfer from English-centric LLMs, which often fail to handle the specific orthography and sociolinguistics of low-resource languages accurately.
\end{itemize}

\section{Discussion and Future Recommendations}
The dual-language text detoxification pipeline demonstrates quality model performance for isiXhosa and Yorùbá. The Logistic Regression classifiers achieved accuracies ranging from 61\% to 72\% for isiXhosa and 72\% to 86\% for Yorùbá, with ROC-AUC values up to 0.88, indicating strong discriminatory capability. Precision, recall, and F1-score metrics further demonstrated that the models reliably identify toxic sentences while maintaining low false positive rates for non-toxic inputs. Detailed per-fold confusion matrices are presented in Figures \ref{fig:kfold_conf_xh} and \ref{fig:kfold_conf_yo}, highlighting consistent detection performance across folds for both languages.

The feature importance visualisations (Figures \ref{fig:weight_xh} and \ref{fig:weight_yo}) show the top TF--IDF tokens contributing to toxicity classification. These plots enhance interpretability and provide insight into which lexical items influence model predictions, supporting the lexicon-guided rewriting component. The aggregated metrics across folds, summarised in Table \ref{tab:aggregated_results}, confirm overall detection performance. The results indicate that Yorùbá benefits from higher accuracy and F1-scores, likely due to a more balanced dataset. In contrast, isiXhosa achieves moderate yet reliable detection, with token-level replacements providing additional interpretability and semantic preservation.

The GUI-based demonstrations provided qualitative validation of the pipeline. As illustrated in Figure \ref{fig:gui_detox}, input sentences were correctly classified as `[TOXIC]` or `[NON-TOXIC]`, with toxic sentences subsequently rewritten via full-sentence lookup or token-level replacement. Non-toxic sentences remained unchanged. Observations confirm that the detoxified outputs retain semantic content while replacing offensive expressions, and that the interface provides an intuitive tool for native speakers to assess cultural and linguistic appropriateness.

Together, these results demonstrate that the dual-language pipeline achieves robust toxic detection and effective meaning-preserving detoxification, offering a computationally efficient and interpretable solution suitable for low-resource African language contexts.

Despite these encouraging outcomes, some limitations remain. The reliance on keyword-driven labelling, though validated by native speakers, may overlook implicit toxicity or contextually nuanced expressions. Additionally, the binary classification formulation does not capture degrees or types of toxicity, which are important for more fine-grained moderation systems.

\subsection{Future Recommendations}
Based on the results and limitations observed in this study, the following recommendations are proposed for future work, with consideration for low-resource environments:

\begin{itemize}
	\item The curated parallel datasets for isiXhosa and Yorùbá should be expanded in size and diversity to enhance the robustness of both toxicity detection and meaning-preserving rewriting, particularly for idiomatic and culturally specific expressions. 	
	\item The integration of multilingual sequence-to-sequence models, such as mT5 and Flan-T5, could be explored to improve the handling of context-dependent and nuanced toxic expressions while maintaining semantic fidelity; however, parameter-efficient tuning strategies (e.g., adapters \cite{Houlsby2019}, LoRA \cite{hu2022}, or prompt-tuning \cite{lester2021,han2024}) should be employed to reduce computational overhead suitable for low-resource settings.
	
	\item Hybrid approaches that combine lexicon-guided methods with lightweight neural generative models could be developed to enable more flexible and context-aware detoxification while retaining interpretability.
	
	\item The incorporation of fine-grained text categories detection, such as insults, threats, or harassment, can be explored to provide a more detailed analysis of toxic content. This would enable targeted detoxification strategies and a more comprehensive evaluation of language-specific toxicity patterns. 
\end{itemize}

\section{Conclusion}
This study presents a dual-language text detoxification pipeline for isiXhosa and Yorùbá, combining TF--IDF-based feature extraction, logistic regression detection, and meaning-preserving rewriting through dataset lookup and token-level replacement. The approach achieves robust toxicity detection, interpretable feature importance, and culturally appropriate detoxified outputs for low-resource settings. The GUI-based demonstrations and quantitative evaluations confirm the pipeline's effectiveness, while language-specific analysis highlights the unique challenges of each language. The results underscore the feasibility of lightweight, interpretable models for low-resource African languages, providing a foundation for the potential development of multilingual generative models and expanded datasets as future research.

\section*{Data Availability}

The curated datasets used in this study, including the parallel toxic $\rightarrow$ detoxified sentence pairs for isiXhosa and Yorùbá, are publicly available through Mendeley Data (DOI: https://doi.org/10.17632/jz8mpwdmgr.1).

\section*{Funding}
The author declare that no funding was received from any organization or agency in support of this research.

%
%
\bibliographystyle{unsrt}
\bibliography{mybibliography}
\end{document}